# Three-dimensional Torques and Power of Horse Forelimb Joints at Trot

H. M. Clayton, D. H. Sha, and D. R. Mullineaux

McPhail Equine Performance Center, Michigan State University
East Lansing, MI 48824, USA, Tel: 517 432 5927, Fax: 517 353 7733
Email: claytonh@cvm.msu.edu

## Summary

**REASONS FOR PERFORMING STUDY:** Equine gait analysis has focused on 2D analysis in the sagittal plane, while descriptions of 3D kinetics and ground reaction force could provide more information on the Equine gait analysis.
**HYPOTHESIS OR OBJECTIVES:** The aim of this study was to characterize the 3D torques and powers of the forelimb joints at trotting.
**METHODS:** Eight sound horses were used in the study. A full 3D torque and power for elbow, carpus, fetlock, pastern and coffin joints of right forelimb in horses at trot were obtained by calculating the inverse kinetics of simplified link segmental model.
**RESULTS:** Over two third of energy (70%) generated by all joints come from stance phase, and most of energy generated was by elbow joint both in stance (77%) and sway (88%) phases. Energy absorbed by all joints during stance (40%) and sway (60%) phases respectively is not a big difference. During stance phase, all most two third of energy (65%) absorbed was by fetlock joint, while over two third of energy (74%) absorbed was by carpus joint during sway phase.
**CONCLUSIONS & CLINICAL RELEVANCE:** This study presents a full 3D kinetic analysis of the relative motion of the humerus, radius, cannon, pastern and coffin segments of the forelimb at the trot. The results could provide for a more sensitive measure for kinetic analysis.

## Introduction

Kinematics refers to the positions of the various body segments and joints as they move with time, including their linear and angular velocities. Kinetics refers to the forces, moments (torques) and powers responsible for the motion. Although the kinematic changes of limbs during motion can be used to visually assess the gait of horses by a clinician, the joint torques/moments and joint powers related to the kinetic changes can provide the clinician even more information on this subject (Clayton *et al.* 2000) because these kinetic variables describe what is happening in the muscles to cause the joint motions observed. These joint torques and joint powers can be obtained from the kinematics by a process known as inverse dynamics, in which a mathematical model of the limb is derived according to the simple laws of mechanics (Newton-Euler equations; Craig 1995).

Using inverse dynamics to calculate the net joint moments and joint power from the knowledge of the kinematics, ground reaction forces, morphological data, and a link-segment model have been performed by several researchers (Schryver *et al*. 1978; Hjerten and Drevemo 1987; Clayton *et al*. 1998; 2000; Colborne *et al*. 1998; Lanovaz *et al*. 1999; Lanovaz *et al*. 2001). The complex structure of the limbs is simplified by representing the limb segments as rigid bodies linked by simple hinge joints. These studies assumed that the model is purely two dimensional with the segments of the horse moving in the sagittal plane. The joint torques and joint powers have been used for evaluation of mechanics and energetics of lameness in horses and to measure effects of superficial digital flexor tendonitis (Clayton *et al.* 2000). Abnormal gaits are inefficient and are associated with an overall increase in energy expenditure. Specific gait abnormalities can cause substantially characteristic alternations in the shape of the moment and power profiles as well as changes in the amount of energy absorbed and generated at the joints.

In fact, all motions that occur in most anatomical joints involve three-dimensional (3D) movement which is described by six independent coordinates or degrees of freedom, i.e. three rotations and three translations (Craig 1995). A complete understanding of 3D joint moments and powers is important in the diagnosis of joint disorder resulting from injury or disease (Clayton *et al*. 2000), in the quantitative assessment of treatment, and in the general study of locomotion (Clayton *et al*. 1998).

The primary objective of the study reported here was to provide a full three-dimensional inverse kinetic analysis of horse forelimbs at trot, which has not been investigated previously. And the second objective was to testify the hypothesis that if the kinetics outside the sagittal plane was significant large compared with that within the sagittal plane.

**Materials and Methods**

**Horses** With approval of the institute's Committee on Animal Use and Care, four sound horses were used in the study. The physical data for the subjects are as follows: mass is 433±63kg; length of radius, cannon, pastern and hoof are 369±13mm, 282±16mm, 114±9mm, and 80±5mm, respectively; speed is 3.13±0.15m/s; stride duration was 706±16ms; and stance is 43.5±2.4% of total stride duration.

**Data Collection** To record the kinematics of forelimb of horses, three skin markers were used for the humerus and pastern segments, while bone pin triad markers were used for the radius and cannon segments. The horses were led at the trot along a runway. Three-dimensional trajectories of the markers were recorded at 120 Hz using a video-camera analysis system[1]. At the same time, a 60×120cm force platform[2] embedded in the middle of the walkway was used to collect ground reaction force (GRF) data. Forces in three directions and location of center of pressure were recorded for each trial with a sampling frequency of 1000Hz. The joint kinematics were calculated in terms of helical angles (Grood *et al*. 1983; Woltring 1994) between two segments by using a singular-value decomposition method (Soderkvist and Wedin 1993).

**Model of Forelimb** The right fore limb was simplified as a link model with four segments, i.e. radius, cannon, pastern and hoof. Each joint, i.e. elbow, carpus, fetlock and coffin, has six degrees of freedom (DoF), i.e. three translations and three rotations. Note that the local co-ordinary system (LCS) for limb (Fig. 1a) and model (Fig. 1b) are different with the directions except for the elbow joint. The model with total 24 DoF was built by using the Robotics Toolbox for Matlab. In the LCS for limb: x indicates the cranial (+)/caudal (-) and abduction (+)/adduction (-) axis; y indicates the medial (+)/lateral (-) and flexion (+)/extension (-) axis; z indicates the proximal (+)/distal (-) and internal (+)/external (-) rotation axis. The arrow points towards the positive direction of translations and the rotation direction follows the right-hand rule (except for elbow joint's flexion (-)/extension (+)). In the LCS for model, the kinematic variables outside of brackets, such as $s_1, s_2, s_3, \ldots$, represent translational motion, the arrow points the positive direction of the motion; the variables inside the brackets, such as $r_1, r_2, r_3, \ldots$, represent the rotational motion. The symbol $\otimes$, end of arrow, indicates the direction of general rotational motions in the LCS for model. Except for the internal/external rotation of elbow joint and abduction/adduction for other three joints, the directions of all other rotational motions follow the right-hand rule. Before and after performing the calculation of inverse kinetics, some kinematic and kinetic variable signs need to be changed. The results plotted in the figures were in the LCS for limb (i.e. Fig. 1a).

**Calculation of Net Joint Torques and Power** The segmental parameters, such as mass location of center of mass, and inertia were determined from published data (Buchner *et al*. 1997). The segmental data were combined with joint position (Fig. 1), velocity (first order derivative of position), acceleration (second order derivative of position) and GRF to obtain the joint torque/moment and contact force for elbow, carpus, fetlock and coffin joints via recursive Newton-Euler method by using Robotics Toolbox for Matlab[3]. The joint power is calculated as the product of joint torques and the joint's angular velocity. All variables were expressed in the distal segment anatomical fixed coordinate systems, i.e. in LCS. The consumption of energy during the motion is equal to the work done by horses during the motion, and it can be computed by integrating the power over the time span of motion. The power can be computed as the product of joint torques and the joint's angular velocity. If the joint torque acts in the same direction as joint angular velocity, the power is positive which means energy generation and the muscles perform positive work in which the muscle shortens during muscular contraction (concentric). If joint torque acts in the opposite direction to joint angular velocity, the power is negative which means energy absorption, and the muscles perform negative work in which the muscle lengthens (eccentric) as it generates tension. Energy absorption occurs when the muscles control joint motion in opposition to the influence of gravity or some other external forces. The work performed (energy expended) over a period of time by the muscles and tendons that cross a specific joint is calculated by integration of the power curve with time. This yields the energy absorbed and the energy generated (Clayton *et al.* 2000). All measurements were normalized by the mass of horses.

**Statistical Analyses** Mean values for joint position, velocity, and acceleration were calculated for all horses (n=4). Mean values for GRF (n=5 at least) were calculated for each horse. Data sets were time normalized by piecewise cubic spline interpolation. The power curve was integrated to determine the mechanical energy generated and absorbed at each joint. Total energy generated and total energy absorbed at each joint was determined by summation. Group mean values and s.d. for each measure were calculated for the four horses.

**Results**

The mean (n=4) and s.d. of the range of torque and power and energy/work during stance and swing phases were calculated in the LCS of joints. The curves of mean and s.d. of joint moments, contact forces and powers of joints were plotted in Fig. 3 and 4. The mean and s.d. of joint torques (moments) and contact forces at special points (maximum and minimum) during stance phase are listed in Tables 1 and 2. Compared with other two rotations, the torque for internal/external rotation was small for every joint. But the torque for the adduction/abduction was large relative to the torque for the flexion/extension and even larger than that of the coffin and elbow joints.

From Fig.3, it is clear that the large contact forces occurred during the stance. As were known, the contact forces of joints during swing phase are due to the weights and inertial loads of segments while the contact forces during stance phase are not only dependent on the weights and inertial load of the body and segments but also on the ground forces. The contact forces of joints are dependent on the ground forces during horse trotting by comparing the contact forces during stance phase and swing phase. Due to the vertical contact force being mainly dependent on the weight, the vertical contact force of each joint was in the same level in magnitude and similar patterns as the vertical ground force only with the direction being opposite. Although there was a small ground force in medial/lateral direction, there was a relative large contact force in medial direction for each joint due to the abduction/adduction and internal/external rotation of the joints. The cranial/caudal contact forces of joints were generally small in magnitude except for the coffin joint. All cranial/caudal contact forces of carpus, fetlock and coffin increased from zero at the beginning of the stance phase to the maximum value then decreased to zeros at the end of stance phase. While the cranial/caudal contact force of elbow increased from zero at the beginning of the stance phase to the maximum value at about 20% of stance and decreased a little bit in magnitude kept the value to the end of stance phase. It is the contact force of elbow in the cranial/caudal direction that keeps the body of the horse moving forward.

The energy expenditure and the net energy by each joint are shown in Table 3. During stance phase, elbow and coffin joints generate energy while the carpus and fetlock joins absorb energy. Whereas during swing phase, except for elbow joint the other three joints absorb energy. The energy contribution by each joint was shown in Table 4. Table 4 shows that the energy generated by all joints during stance and swing phases are about same, i.e. 53%±3% and 47%±3% of the total energy generated by all joints during a full stride respectively. During the stance phase, the most energy generated was generated by the elbow, which is about 60%±2% of the total energy generated by all joints. The second most was by fetlock and it is about 30%±2% of the total energy generated by all joints. The least was by coffin and it is only about 2% of the total energy generated by all joints. The energy generated by carpus during stance phase is about 8% of the total energy generated by all joints during stance phase. The most energy generated during swing phase was by elbow, and it is about 88% of the total energy generated by all joints during swing phase. While the other three joints generated the same amount energy by carpus, fetlock and coffin with 6%, 3% and 3% of the total energy generated by all joints during swing phase.

Table 4 also shows that the energy absorbed by all joints during stance phase and swing phase are about same amount in quantity. They are about 48%±2% and 52%±2% of the total energy absorbed by all joints over a full stride. During stance phase, the most energy absorbed was by fetlock, which is about 58%±1% of the total energy absorbed by all joints. Elbow and carpus absorbed the same amount energy that is about 18% of the total energy absorbed by all joints. Only about 6%±2% of the total energy absorbed by all joints was by coffin joint. During swing phase, carpus absorbed the most energy that is about 71% of the total energy absorbed by all joints. Elbow, fetlock and coffin absorbed 9%, 18% and 3% of the total energy absorbed by all joints respectively.

## Discussion

A full 3D kinetic analysis to the elbow, carpus, fetlock and coffin joints of the right forelimb of sound horses at trot was performed. The results presented here are obtained from a fairly uniform population of horses and the findings may vary somewhat in horses with a different conformation and size.

The musculature that drives the limb has a complex geometry and many mechanical and material properties that are difficult to quantify. In order to study the effect of muscle action without defining all of the specific parameters of the system, the net muscular action within the limb is represented in the model as net torques or net joint moments acting at each of the joints. Realistic inertial parameters such as mass and moment of inertia are assigned to each segment of the model. Kinematic data and GRFs are measured and used as inputs to the model. An inverse dynamics solution is used to calculate the torques required to cause the observed motion and forces. This net effect of muscle action derived from the model can be applied in both clinical and research settings to study the function and dysfunction of the limb at the different joints (Buchner *et al*. 1996; Clayton *et al*. 1998). Additional information can be added to the model to increase the level of sophistication. For example, knowledge of the orientation of anatomic structures such as the line of action of tendons can allow for a more detailed estimation of internal muscle and tendon forced (Riemersma *et al*. 1988; Jansen *et al*. 1993).

Net joint moments indicate the summation of all muscle activity and do not distinguish between contributions from different muscle groups. It is prudent to avoid over interpretation of the net values unless electromyography data are collected simultaneously with kinematic and GRF data. Many of the muscular and tendinous structures responsible for moving and stabilizing the equine forelimb segments cross more than one joint. Therefore, complex relationships exist between the movements and moments at different joints (Clayton *et al*. 1998).

In the analysis of energy expenditure, the relationship between mechanical and metabolic energy should be considered (Clayton *et al*. 2000). Concentric muscle contraction leads in generation of mechanical energy, whereas eccentric muscle contraction leads in absorption of mechanical energy. Concentric and eccentric muscle contractions use metabolic energy.

In summary, a full 3D kinetic analysis to the joints: elbow, carpus, fetlock and coffin of right front limbs of sound horses at trot have been reported in this paper. The kinetics outside the sagittal plane was significant large and could not be ignored compared with that within the sagittal plane. The results could provide for a more sensitive measure for kinetic analysis.

## Acknowledgements

We would like to acknowledge the endowment of Mary Anne McPhail for financial support of this work.

## Manufacturers' addresses

[1]Motion Analysis Corporation, Santa Rosa, CA, USA.
[2]AMTI, Watertown, MA, USA.
[3]MathWorks Inc, Natick, MA, USA.

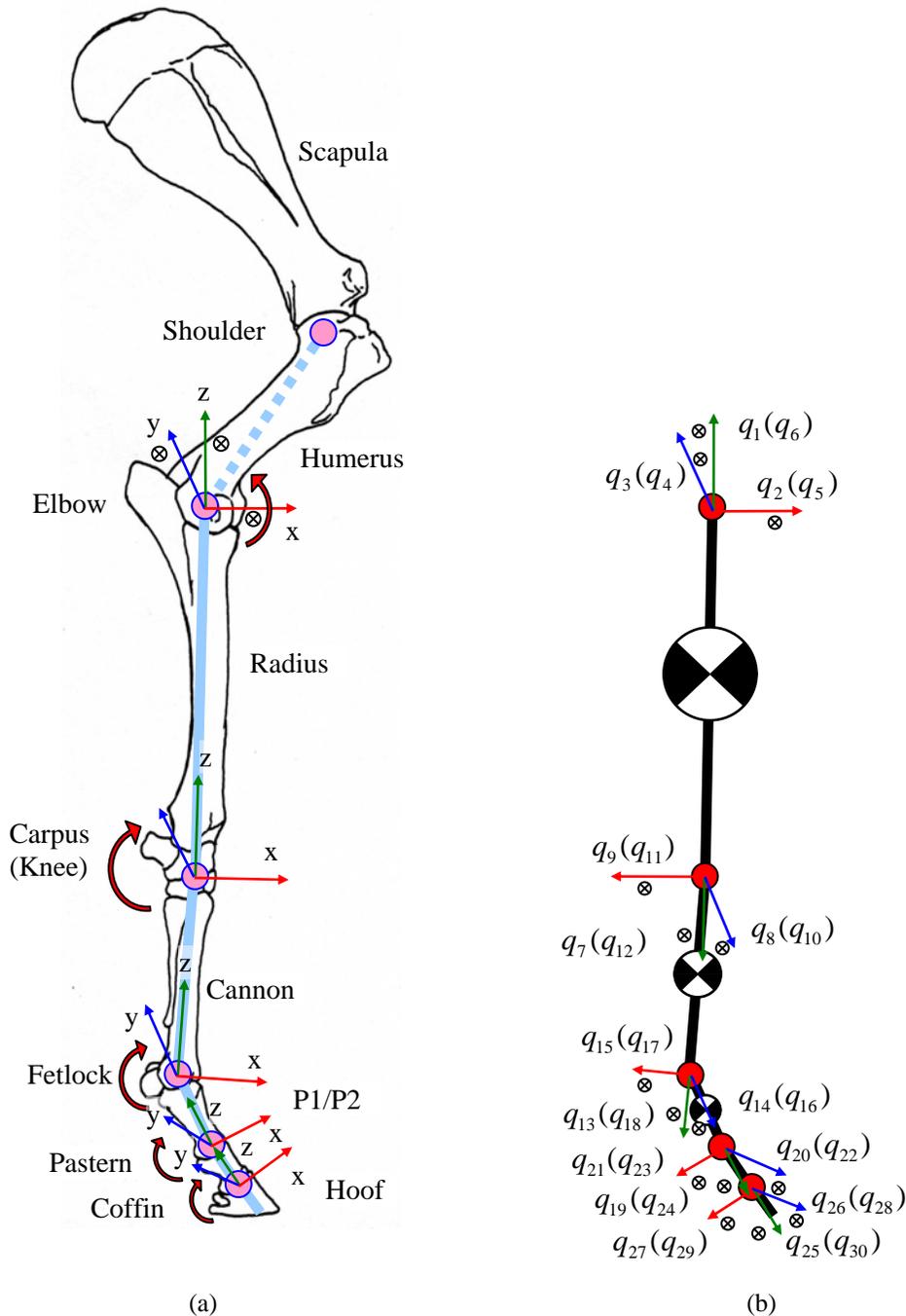

(a)             (b)

**Fig. 1** Link model for forelimb of horses. (a) Forelimb segments, link segmental model and the anatomic LCS for each joint: x- cranial (+)/caudal (-) and adduction (+)/abduction (-) axis; y- medial (+)/lateral (-) and flexion (+)/extension (-) axis; z- proximal (+)/distal (-) and internal (+)/external (-) rotation axis. (b) Computer model and the LCS for each joint: the kinematic variables outside of brackets, such as $q_1, q_2, q_3, \ldots$, represent translational motion, the arrow points the positive direction of the motion; the variables inside the brackets, such as $q_4, q_5, q_6, \ldots$, represent the rotational motion. Except for the internal/external rotation of elbow joint and abduction/adduction for other three joints, the directions of all other rotational motions follow the right-hand rule.

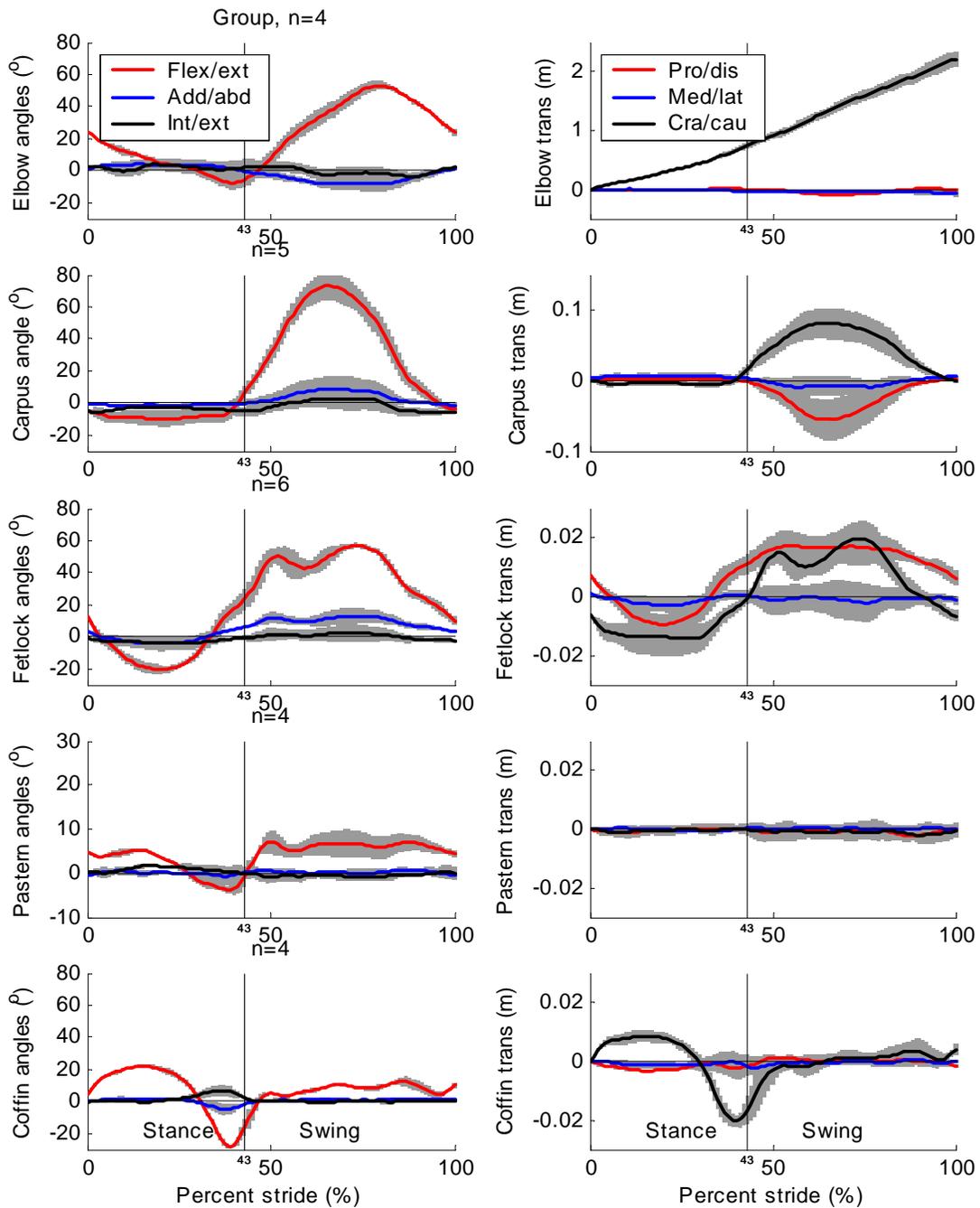

**Fig. 2** Mean joint angles (left) and translations (right): the joint angles were measured relative to the position in which the horse stands squarely. Flexion/adduction/internal-rotations are positive and extension/abduction/external-rotations are negative. The cranial/medial/proximal translations are positive and caudal/lateral/distal translations are negative. Second vertical line indicates the transition from stance phase to swing phase. Translations for joints of carpus, fetlock and coffin were not used in the calculation of inverse kinetics because of their lack of reliability.

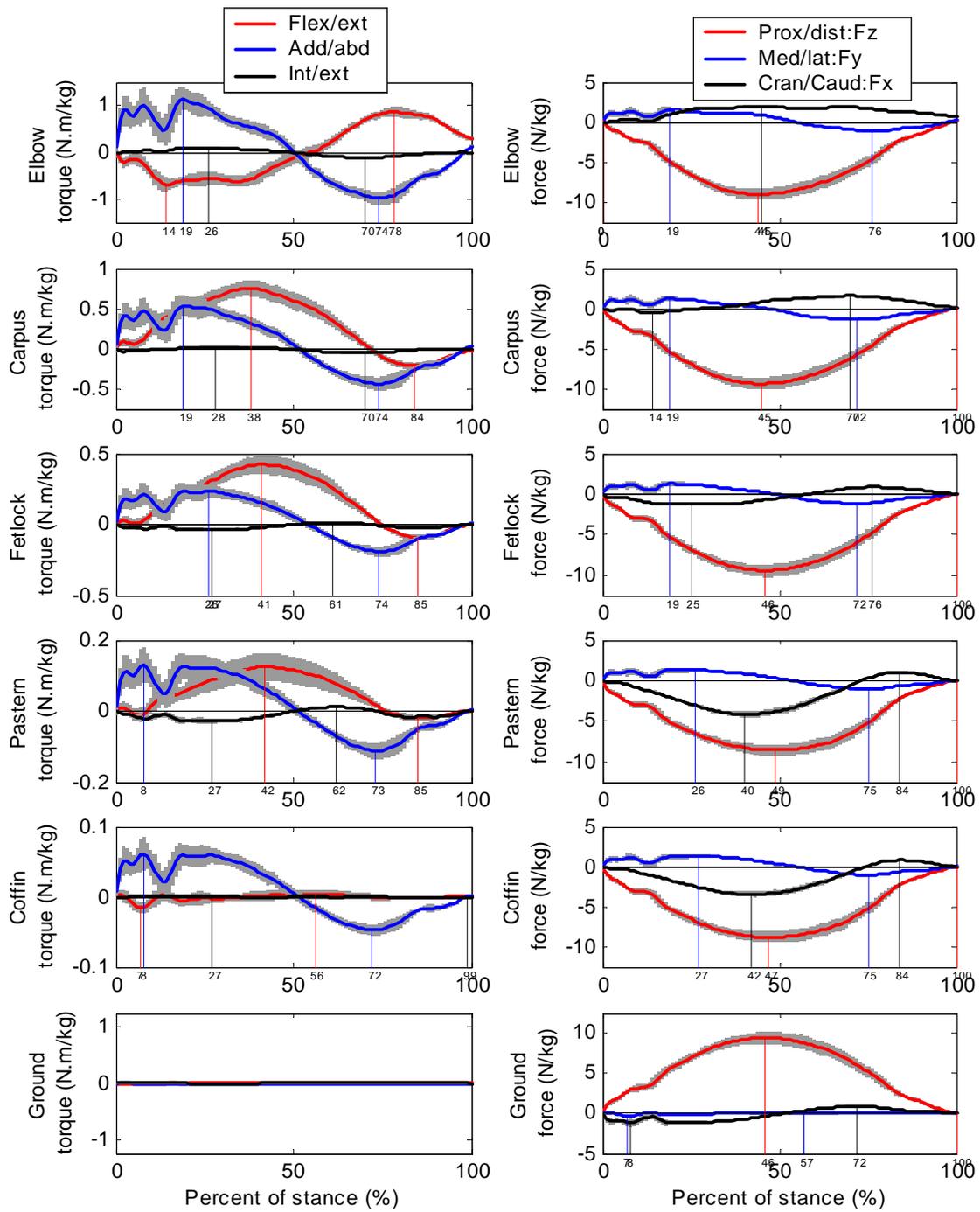

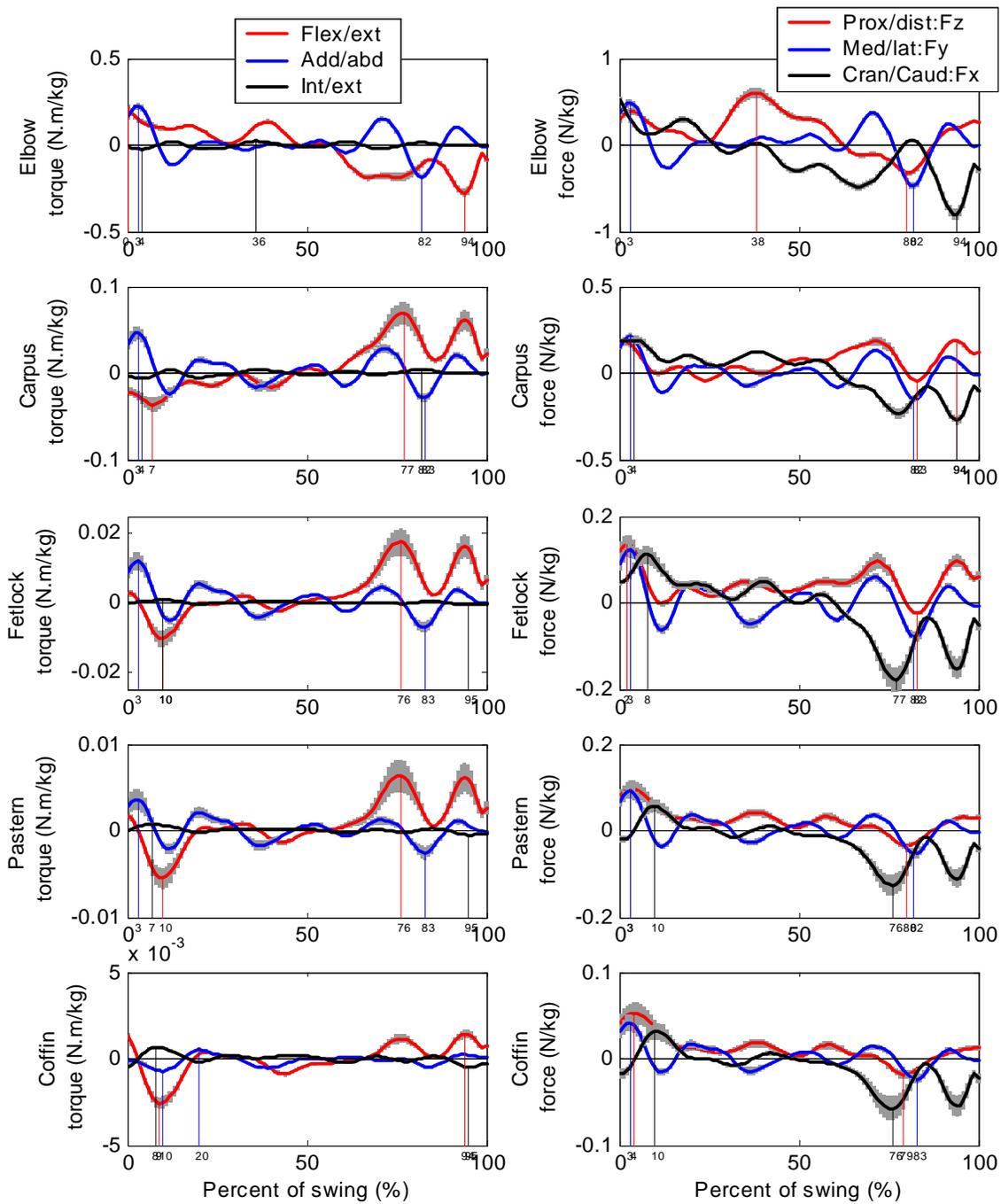

**Fig. 3** Joint moments (torques) and contact forces: left column is the moment for the ground and each joints; right column is the internal contact force for each joints and ground reaction force. First vertical line indicates the change of horizontal ground force in cranial/caudal direction. Second vertical line indicates the transition from the stance phase to the swing phase.

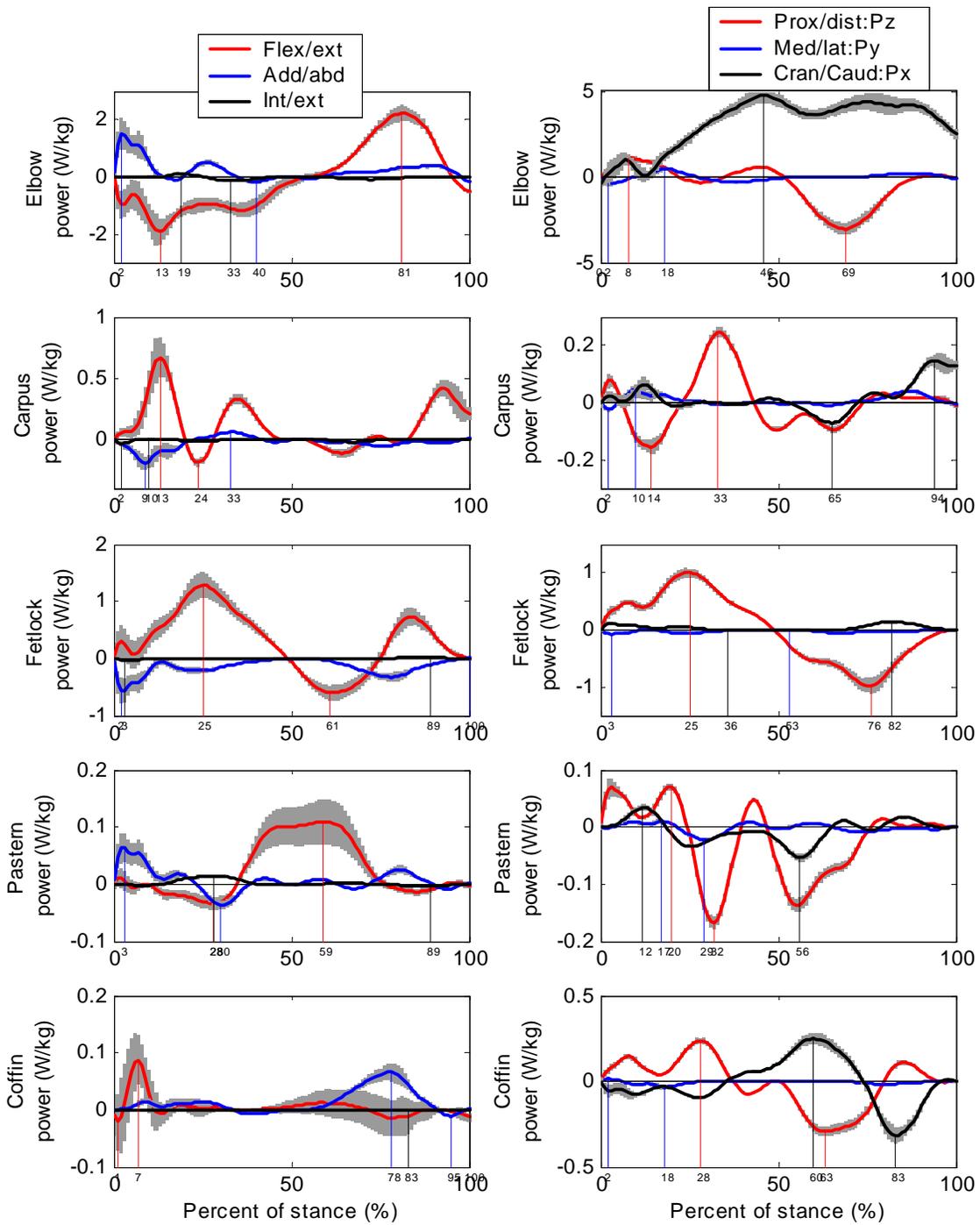
10

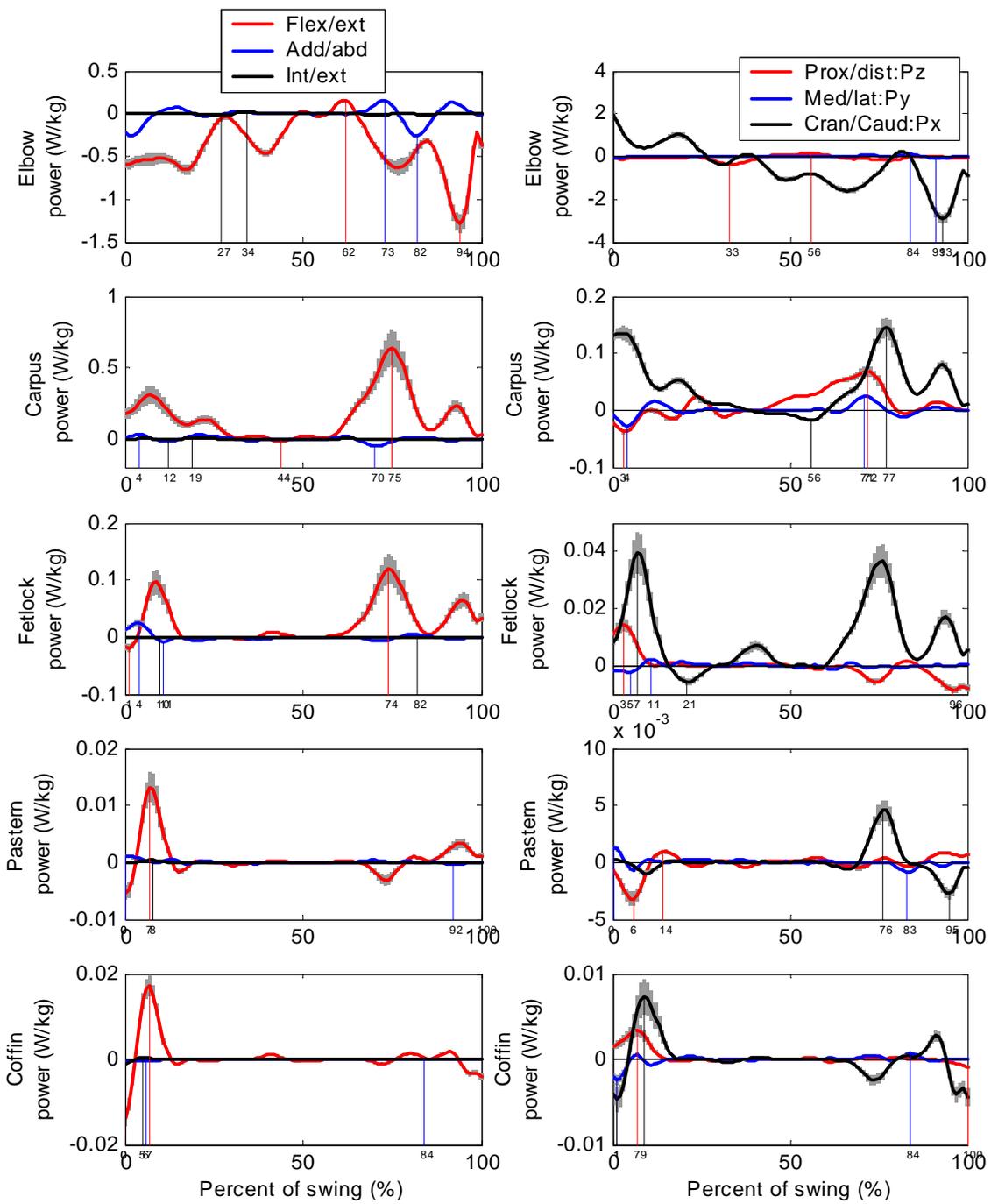

**Fig. 4** Joint powers (n=4): left column is the power of rotational motion for each joint; right column is the power of translation for each joint. Since no translations for joint carpus, fetlock and coffin, no translation power for these joints.

**Table 1**: Physical data for all subjects: The descriptive data are from at least five kinematic trials for each subject. The length of each segment was calculated by two lateral markers at standing position.

For carpus

| Subject | Height at withers (cm) | Mass (kg) | Radius length (mm) | Cannon length (mm) | Avg. speed (m/s) | Stride time (ms) | Stance % of stride |
|---|---|---|---|---|---|---|---|
| Asics | - | 450 | 363 | 129 | 3.12±0.20 | 737±27 | 41.4±1.3 |
| Reebok | 149 | 485 | 382 | 146 | 2.96±0.08 | 703±11 | 45.3±1.1 |
| Speck | 144 | 364 | 369 | 161 | 3.18±0.21 | 723±21 | 44.5±1.3 |
| Han Solo | 151 | 452 | 356 | - | 3.24±0.18 | 691±19 | 40.8±1.6 |
| Mean ± SD | 148±4 | 438±52 | 368±11 | 145±16 | 3.13±0.15 | 706±16 | 43.5±2.4 |

For fetlock

| Subject | Height at withers (cm) | Mass (kg) | Cannon length (mm) | P1 length (mm) | Avg. speed (m/s) | Stride time (ms) | Stance % of stride |
|---|---|---|---|---|---|---|---|
| Asics | - | 450 | 129 | 72 | 3.12±0.20 | 737±27 | 42.3±1.3 |
| Speck | 144 | 364 | 161 | - | 3.14±0.18 | 722±19 | 44.6±1.3 |
| Rizzo | 165 | 544 | 168 | 49 | 2.73±0.14 | 802±24 | 45.8±1.1 |
| Janis | 158 | 501 | 184 | 45 | 2.95±0.09 | 738±11 | 46.6±1.1 |
| Pat | 155 | 533 | 190 | 46 | 3.03±0.08 | 760±22 | 43.8±0.8 |
| Jimi | 164 | 508 | 195 | 54 | 3.37±0.11 | 783±26 | 41.9±1.1 |
| Mean ± SD | 157±8 | 483±67 | 171±24 | 56±10 | 3.05±0.21 | 754±29 | 44.3±1.7 |

For pastern

| Subject | Height at withers (cm) | Mass (kg) | Avg. speed (m/s) | Stride time (ms) | Stance % of stride |
|---|---|---|---|---|---|
| Rizzo | 165 | 544 | 2.73±0.14 | 802±24 | 45.8±1.1 |
| Janis | 158 | 501 | 2.95±0.09 | 738±11 | 46.6±1.1 |
| Pat | 155 | 533 | 3.09±0.13 | 768±21 | 44.3±1.1 |
| Jimi | 164 | 508 | 3.58±0.10 | 744±14 | 42.5±0.8 |
| Mean ± SD | 160±5 | 521±20 | 3.09±0.36 | 763±29 | 44.8±1.8 |

For coffin

| Subject | Height at withers (cm) | Mass (kg) | Avg. speed (m/s) | Stride time (ms) | Stance % of stride |
|---|---|---|---|---|---|
| Rizzo | 165 | 544 | 2.81±0.08 | 798±10 | 44.8±1.9 |
| Janis | 158 | 501 | 2.95±0.26 | 731±16 | 48.4±3.2 |
| Pat | 155 | 533 | 3.09±0.13 | 768±21 | 44.3±1.1 |
| Jimi | 164 | 508 | 3.59±0.10 | 744±14 | 42.5±0.8 |
| Mean ± SD | 161±5 | 522±20 | 3.11±0.34 | 760±30 | 45±2.5 |

**Table 2** The kinematics of joints used in the inverse kinetics for each horse. Letter "Y" means the kinematics of joint was used.

| Horse | Joints | | | | |
|---|---|---|---|---|---|
| | Elbow | Carpus | Fetlock | Pastern | Coffin |
| Asics | Y | Y | Y | | |
| Reebok | Y | Y | | | |
| Speck | Y | Y | Y | | |
| Han Solo | Y | | | | |
| Rizzo | | Y | Y | Y | Y |
| Janis | | Y | Y | Y | Y |
| Pat | | | Y | Y | Y |
| Jimi | | | Y | Y | Y |

**Table 3** the relationships between anatomical convention and the convention for general co-ordinates of link model

| Joint | Anatomic Motion | Symbol of Kinematics | General co-ordinates of Limb Model | Sign of Torque (T) | Sign of Power (P) |
|---|---|---|---|---|---|
| Elbow | Proximal/distal | Z | $q_1 = z$ | + | + |
| | Medial/lateral | Y | $q_3 = y$ | + | + |
| | Cranial/caudal | X | $q_2 = x$ | + | + |
| | Flexion/extension | β | $q_4 = -\beta$ | - | - |
| | Adduction/abduction | α | $q_5 = \alpha$ | + | + |
| | Internal/external | γ | $q_6 = -\gamma$ | - | - |
| Carpus | Proximal/distal | Z | $q_7 = -z$ | - | - |
| | Medial/lateral | Y | $q_8 = -y$ | - | - |
| | Cranial/caudal | X | $q_9 = -x$ | - | - |
| | Flexion/extension | β | $q_{10} = -\beta$ | - | - |
| | Adduction/abduction | α | $q_{11} = \alpha$ | + | + |
| | Internal/external | γ | $q_{12} = -\gamma$ | - | - |
| Fetlock | Proximal/distal | Z | $q_{13} = -z$ | - | - |
| | Medial/lateral | Y | $q_{14} = -y$ | - | - |
| | Cranial/caudal | X | $q_{15} = -x$ | - | - |
| | Flexion/extension | β | $q_{16} = -\beta$ | - | - |
| | Adduction/abduction | α | $q_{17} = \alpha$ | + | + |
| | Internal/external | γ | $q_{18} = -\gamma$ | - | - |
| Pastern | Proximal/distal | Z | $q_{19} = -z$ | - | - |
| | Medial/lateral | Y | $q_{20} = -y$ | - | - |
| | Cranial/caudal | X | $q_{21} = -x$ | - | - |
| | Flexion/extension | β | $q_{22} = -\beta$ | - | - |
| | Adduction/abduction | α | $q_{23} = \alpha$ | + | + |
| | Internal/external | γ | $q_{24} = -\gamma$ | - | - |
| Coffin | Proximal/distal | Z | $q_{25} = -z$ | - | - |
| | Medial/lateral | Y | $q_{26} = -y$ | - | - |
| | Cranial/caudal | X | $q_{27} = -x$ | - | - |
| | Flexion/extension | β | $q_{28} = -\beta$ | - | - |
| | Adduction/abduction | α | $q_{29} = \alpha$ | + | + |
| | Internal/external | γ | $q_{30} = -\gamma$ | - | - |

**Table 4** Maximum and minimum joint torques, mean (s.d.), and the time of occurrence during stance phase of horses trotting at 3.13±0.15m/s. The stance phase ends at the time of 43% of a full stride. The values were normalized to the body mass.

| Joint | Rotation | Normal torque (N.m/kg) | |
|---|---|---|---|
| | | Max. (s.d.) at % of stance | Min. (s.d.) at % of stance |
| Elbow | Flex/ext | 0.8597(0.0956) at 78% | -0.6863(0.1437) at 14% |
| | Add/abd | 1.1361(0.0956) at 19% | -0.9808(0.1408) at 74% |
| | Int/ext | 0.0946(0.0956) at 26% | -0.1116(0.0183) at 70% |
| Carpus | Flex/ext | 0.7570(0.0953) at 38% | -0.2102(0.0363) at 84% |
| | Add/abd | 0.5360(0.0953) at 19% | -0.4351(0.0816) at 74% |
| | Int/ext | 0.0254(0.0953) at 28% | -0.0527(0.0079) at 70% |
| Fetlock | Flex/ext | 0.4229(0.0633) at 41% | -0.0876(0.0165) at 85% |
| | Add/abd | 0.2348(0.0633) at 26% | -0.1898(0.0341) at 74% |
| | Int/ext | 0.0145(0.0633) at 61% | -0.0384(0.0074) at 27% |
| Pastern | Flex/ext | 0.1242(0.0382) at 42% | -0.0200(0.0083) at 85% |
| | Add/abd | 0.1288(0.0382) at 8% | -0.1118(0.0223) at 73% |
| | Int/ext | 0.0119(0.0382) at 62% | -0.0291(0.0054) at 27% |
| Coffin | Flex/ext | 0.0041(0.0069) at 56% | -0.0158(0.0079) at 7% |
| | Add/abd | 0.0611(0.0069) at 8% | -0.0469(0.0079) at 72% |
| | Int/ext | 0.0002(0.0069) at 27% | -0.0006(0.0001) at 99% |

**Table 5** Maximum and minimum joint forces, mean (s.d.), and the time of occurrence during stance phase. The stance phase ends at the time of 43% of a full stride. The values were normalized to the body mass.

| Joint | Translation | Normal internal force (N/kg) | |
|---|---|---|---|
| | | Max. (s.d.) at % of stance | Min. (s.d.) at % of stance |
| Elbow | Prox/dist | 0.2875(0.3253) at 0% | -9.0029(0.7039) at 44% |
| | Medio/lateral | 1.5607(0.3253) at 19% | -1.0904(0.1465) at 76% |
| | Cranial/caudal | 1.9216(0.3253) at 45% | -0.1238(0.0662) at 0% |
| Carpus | Prox/dist | 0.1050(0.1043) at 100% | -9.3510(0.7429) at 45% |
| | Medio/lateral | 1.2724(0.1043) at 19% | -1.3270(0.1596) at 72% |
| | Cranial/caudal | 1.6235(0.1043) at 70% | -0.5613(0.1585) at 14% |
| Fetlock | Prox/dist | 0.0636(0.1041) at 100% | -9.3948(0.7504) at 46% |
| | Medio/lateral | 1.2900(0.1041) at 19% | -1.1502(0.1452) at 72% |
| | Cranial/caudal | 0.8873(0.1041) at 76% | -1.2754(0.1501) at 25% |
| Pastern | Prox/dist | 0.0394(0.0945) at 100% | -8.5647(0.7025) at 49% |
| | Medio/lateral | 1.3499(0.0945) at 26% | -1.0175(0.1489) at 75% |
| | Cranial/caudal | 0.9798(0.0945) at 84% | -4.1581(0.3262) at 40% |
| Coffin | Prox/dist | 0.0029(0.0957) at 100% | -8.8027(0.7029) at 47% |
| | Medio/lateral | 1.3427(0.0957) at 27% | -1.0031(0.1467) at 75% |
| | Cranial/caudal | 0.8316(0.0957) at 84% | -3.4439(0.3014) at 42% |
| GRF | Vertical | 9.4412(0.7567) at 46% | 0.0332(0.1030) at 100% |
| | Transverse | 0.0862(0.7567) at 57% | -0.3069(0.1612) at 7% |
| | Forward | 0.8962(0.7567) at 72% | -1.1699(0.4546) at 8% |

**Table 6** Max min normal torques to body mass over swing phase

| Joint | Rotation | Normal torque (N.m/kg) | |
| --- | --- | --- | --- |
| | | Max. (s.d.) at % of swing | Min. (s.d.) at % of swing |
| Elbow | Flex/ext | 0.2180(0.0197) at 0% | -0.2759(0.0242) at 94%, |
| | Add/abd | 0.2260(0.0197) at 3% | -0.1829(0.0132) at 82%, |
| | Int/ext | 0.0238(0.0197) at 36% | -0.0235(0.0034) at 4%, |
| Carpus | Flex/ext | 0.0692(0.0137) at 77% | -0.0367(0.0077) at 7%, |
| | Add/abd | 0.0469(0.0137) at 3% | -0.0286(0.0046) at 83%, |
| | Int/ext | 0.0043(0.0137) at 82% | -0.0059(0.0013) at 4%, |
| Fetlock | Flex/ext | 0.0176(0.0039) at 76% | -0.0104(0.0023) at 10%, |
| | Add/abd | 0.0120(0.0039) at 3% | -0.0072(0.0014) at 83%, |
| | Int/ext | 0.0009(0.0039) at 10% | -0.0006(0.0001) at 95%, |
| Pastern | Flex/ext | 0.0064(0.0019) at 76% | -0.0055(0.0012) at 10%, |
| | Add/abd | 0.0036(0.0019) at 3% | -0.0025(0.0007) at 83%, |
| | Int/ext | 0.0008(0.0019) at 7% | -0.0005(0.0001) at 95%, |
| Coffin | Flex/ext | 0.0014(0.0003) at 94% | -0.0026(0.0004) at 9%, |
| | Add/abd | 0.0005(0.0003) at 20% | -0.0007(0.0001) at 10%, |
| | Int/ext | 0.0007(0.0003) at 8% | -0.0005(0.0001) at 95%, |

**Table 7** Max min normal internal force to body mass over swing phase

| Joint | Translation | Normal internal force (N/kg) | |
| --- | --- | --- | --- |
| | | Max. (s.d.) at % of swing | Min. (s.d.) at % of swing |
| Elbow | Prox/dist | 0.6001(0.0576) at 39% | -0.3191(0.0288) at 81% |
| | Medio/lateral | 0.4919(0.0576) at 4% | -0.4726(0.0401) at 83% |
| | Cranial/caudal | 0.5270(0.0576) at 1% | -0.7977(0.0682) at 95% |
| Carpus | Prox/dist | 0.1866(0.0127) at 95% | -0.0436(0.0040) at 84% |
| | Medio/lateral | 0.2131(0.0127) at 4% | -0.1499(0.0091) at 83% |
| | Cranial/caudal | 0.1912(0.0127) at 5% | -0.2722(0.0243) at 95% |
| Fetlock | Prox/dist | 0.1352(0.0230) at 3% | -0.0235(0.0039) at 84% |
| | Medio/lateral | 0.1236(0.0230) at 4% | -0.0794(0.0113) at 83% |
| | Cranial/caudal | 0.1118(0.0230) at 9% | -0.1785(0.0291) at 78% |
| Pastern | Prox/dist | 0.0961(0.0188) at 4% | -0.0348(0.0061) at 81% |
| | Medio/lateral | 0.0917(0.0188) at 4% | -0.0536(0.0091) at 83% |
| | Cranial/caudal | 0.0579(0.0188) at 11% | -0.1253(0.0244) at 77% |
| Coffin | Prox/dist | 0.0529(0.0128) at 5% | -0.0188(0.0042) at 80% |
| | Medio/lateral | 0.0416(0.0128) at 4% | -0.0236(0.0052) at 84% |
| | Cranial/caudal | 0.0323(0.0128) at 11% | -0.0577(0.0139) at 77% |

**Table 8** Max min normal power of rotations to body mass over stance phase

| Joint | Rotation | Normal power (W/kg) | |
| --- | --- | --- | --- |
| | | Max. (s.d.) at % of stance | Min. (s.d.) at % of stance |
| Elbow | Flex/ext | 2.2223(0.2844) at 81% | -1.8950(0.4532) at 13% |
| | Add/abd | 1.4904(0.2844) at 2% | -0.1609(0.0448) at 40% |
| | Int/ext | 0.1145(0.2844) at 19% | -0.1395(0.0276) at 33% |
| Carpus | Flex/ext | 0.6709(0.1618) at 13% | -0.1896(0.0295) at 24% |
| | Add/abd | 0.0588(0.1618) at33% | -0.1976(0.0523) at 9% |
| | Int/ext | 0.0053(0.1618) at 10% | -0.0342(0.0149) at 2% |
| Fetlock | Flex/ext | 1.2825(0.2170) at 25% | -0.5985(0.1298) at 61% |
| | Add/abd | 0.0111(0.2170) at100% | -0.5601(0.2307) at 2% |
| | Int/ext | 0.0132(0.2170) at 91% | -0.0225(0.0083) at 3% |
| Pastern | Flex/ext | 0.0713(0.0061) at 20% | -0.1676(0.0105) at 32% |
| | Add/abd | 0.0105(0.0061) at 17% | -0.0233(0.0030) at 29% |
| | Int/ext | 0.0338(0.0061) at 12% | -0.0538(0.0064) at 56% |
| Coffin | Flex/ext | 0.2375(0.0150) at 28% | -0.2894(0.0294) at 63% |
| | Add/abd | 0.0152(0.0150) at 2% | -0.0328(0.0096) at 19% |
| | Int/ext | 0.2491(0.0150) at 60% | -0.3153(0.0482) at 83% |

**Table 9** Max min normal translation power to body mass over stance phase

| Joint | Translation | Normal Power (W/kg) | |
| --- | --- | --- | --- |
| | | Max. (s.d.) at % of stance | Min. (s.d.) at % of stance |
| Elbow | Prox/dist | 1.0980(0.1328) at 8% | -3.0380(0.3701) at 69% |
| | Medio/lateral | 0.4816(0.1328) at 18% | -0.4095(0.1469) at 2% |
| | Cranial/caudal | 4.7407(0.1328) at 46% | -0.3559(0.1902) at 0% |
| Carpus | Prox/dist | 0.2476(0.0154) at 33% | -0.1544(0.0275) at 14% |
| | Medio/lateral | 0.0437(0.0154) at 10% | -0.0233(0.0091) at 2% |
| | Cranial/caudal | 0.1474(0.0154) at 94% | -0.0712(0.0063) at 65% |
| Fetlock | Prox/dist | 1.0063(0.0714) at 25% | -0.9747(0.1343) at 76% |
| | Medio/lateral | 0.0036(0.0714) at 53% | -0.0719(0.0243) at 3% |
| | Cranial/caudal | 0.1385(0.0714) at 82% | -0.0121(0.0011) at 36% |
| Pastern | Prox/dist | 0.1098(0.0392) at 59% | -0.0341(0.0115) at 28% |
| | Medio/lateral | 0.0632(0.0392) at 3% | -0.0358(0.0077) at 30% |
| | Cranial/caudal | 0.0147(0.0392) at 28% | -0.0043(0.0010) at 89% |
| Coffin | Prox/dist | 0.0875(0.0436) at 7% | -0.0202(0.0490) at 1% |
| | Medio/lateral | 0.0659(0.0436) at 78% | -0.0106(0.0028) at 95% |
| | Cranial/caudal | 0.0003(0.0436) at 83% | -0.0015(0.0002) at100% |

**Table 10** Max min normal power to body mass over swing phase

| Joint | Rotation | Normal power (W/kg) | |
|---|---|---|---|
| | | Max. (s.d.) at % of swing | Min. (s.d.) at % of swing |
| Elbow | Flex/ext | 0.1543(0.0158) at 62% | -1.2876(0.1129) at 94% |
| | Add/abd | 0.1544(0.0158) at 73% | -0.2564(0.0184) at 82% |
| | Int/ext | 0.0285(0.0158) at 34% | -0.0220(0.0040) at 27% |
| Carpus | Flex/ext | 0.6373(0.1232) at 75% | -0.0138(0.0046) at 44% |
| | Add/abd | 0.0302(0.1232) at 4% | -0.0511(0.0086) at 70% |
| | Int/ext | 0.0052(0.1232) at 19% | -0.0057(0.0010) at 12% |
| Fetlock | Flex/ext | 0.1187(0.0259) at 74% | -0.0188(0.0065) at 1% |
| | Add/abd | 0.0243(0.0259) at 4% | -0.0075(0.0015) at 11% |
| | Int/ext | 0.0001(0.0259) at 82% | -0.0008(0.0001) at 10% |
| Pastern | Flex/ext | 0.0132(0.0027) at 7% | -0.0053(0.0013) at 0% |
| | Add/abd | 0.0011(0.0027) at 0% | -0.0005(0.0001) at 92% |
| | Int/ext | 0.0004(0.0027) at 8% | -0.0001(0.0000) at 100% |
| Coffin | Flex/ext | 0.0171(0.0023) at 7% | -0.0137(0.0020) at 0% |
| | Add/abd | 0.0001(0.0023) at 84% | -0.0004(0.0001) at 6% |
| | Int/ext | 0.0004(0.0023) at 5% | -0.0012(0.0002) at 0% |

**Table 11** Max min normal power to body mass over swing phase

| Joint | Translation | Normal Power (W/kg) | |
|---|---|---|---|
| | | Max. (s.d.) at % of swing | Min. (s.d.) at % of swing |
| Elbow | Prox/dist | 0.1351(0.0139) at 56% | -0.3774(0.0375) at 33% |
| | Medio/lateral | 0.1131(0.0139) at 84% | -0.1084(0.0092) at 91% |
| | Cranial/caudal | 1.9490(0.0139) at 0% | -2.9018(0.2461) at 93% |
| Carpus | Prox/dist | 0.0685(0.0089) at 72% | -0.0372(0.0042) at 3% |
| | Medio/lateral | 0.0241(0.0089) at 71% | -0.0277(0.0020) at 4% |
| | Cranial/caudal | 0.1459(0.0089) at 77% | -0.0175(0.0030) at 56% |
| Fetlock | Prox/dist | 0.0144(0.0024) at 3% | -0.0087(0.0013) at 96% |
| | Medio/lateral | 0.0022(0.0024) at 11% | -0.0023(0.0003) at 5% |
| | Cranial/caudal | 0.0399(0.0024) at 7% | -0.0056(0.0011) at 21% |
| Pastern | Prox/dist | 0.0009(0.0002) at 14% | -0.0032(0.0006) at 6% |
| | Medio/lateral | 0.0013(0.0002) at 0% | -0.0009(0.0002) at 83% |
| | Cranial/caudal | 0.0046(0.0002) at 76% | -0.0028(0.0005) at 95% |
| Coffin | Prox/dist | 0.0034(0.0008) at 7% | -0.0009(0.0002) at 100% |
| | Medio/lateral | 0.0007(0.0008) at 84% | -0.0024(0.0005) at 1% |
| | Cranial/caudal | 0.0073(0.0008) at 9% | -0.0047(0.0015) at 1% |

**Table 12** Energy expenditure of joints, mean (s.d.). Positive means energy generation and negative means energy absorption.

| Joint | Energy generated (J/kg) | | Energy absorbed (J/kg) | | Net energy (J/kg) | |
|---|---|---|---|---|---|---|
| | Stance | Swing | Stance | Swing | Stance | Swing |
| Elbow | 1.3325(0.0887) | 0.1234(0.0099) | -0.4154(0.0533) | -0.5207(0.0394) | 0.9171(0.0758) | -0.3973(0.0301) |
| Carpus | 0.0698(0.0049) | 0.0974(0.0147) | -0.0351(0.0047) | -0.0071(0.0008) | 0.0348(0.0033) | 0.0903(0.0139) |
| Fetlock | 0.2266(0.0242) | 0.0171(0.0033) | -0.1664(0.0243) | -0.0021(0.0004) | 0.0603(0.0111) | 0.0150(0.0029) |
| Pastern | 0.0214(0.0047) | 0.0008(0.0002) | -0.0207(0.0021) | -0.0005(0.0001) | 0.0007(0.0027) | 0.0003(0.0001) |
| Coffin | 0.0414(0.0046) | 0.0010(0.0002) | -0.0425(0.0036) | -0.0007(0.0001) | -0.0010(0.0032) | 0.0003(0.0001) |
| Total | 1.6917 | 0.2398 | -0.68 | -0.5312 | 1.0117 | -0.2914 |

**Table 13** Energy expenditure, mean (s.d.). The data in the table are only for the summations of energy of three rotations of each joint.

| Joint | Energy generated: 100% | | Energy absorbed: 100% | |
|---|---|---|---|---|
| | Stance: 81%(1%) | Swing: 19%(1%) | Stance: 60%(3%) | Swing: 40%(3%) |
| Elbow | 54%(3%) | 16%(2%) | 63%(2%) | 97%(0%) |
| Carpus | 11%(1%) | 71%(2%) | 7%(0%) | 2%(0%) |
| Fetlock | 30%(2%) | 12%(2%) | 28%(1%) | 1%(0%) |
| Pastern | 3%(1%) | 1%(0%) | 1%(0%) | 0%(0%) |
| Coffin | 2%(0%) | 1%(0%) | 1%(1%) | 0%(0%) |

**Table 14** Energy expenditure, mean (s.d.). The data in the table are only for the summations of energy of three translations of each joint.

| Joint | Energy generated: 100% | | Energy absorbed: 100% | |
|---|---|---|---|---|
| | Stance: 90%(1%) | Swing: 10%(1%) | Stance: 53%(3%) | Swing: 47%(3%) |
| Elbow | 87(0%) | 77%(1%) | 60%(0%) | 98%(0%) |
| Carpus | 2%(0%) | 19%(1%) | 4%(0%) | 1%(0%) |
| Fetlock | 8%(0%) | 3%(1%) | 22%(0%) | 1%(0%) |
| Pastern | 1%(0%) | 0%(0%) | 4%(0%) | 0%(0%) |
| Coffin | 3%(0%) | 0%(0%) | 10%(0%) | 0%(0%) |

**Table 15** Energy expenditure, mean (s.d.). The data in the table are for the summations of energy of three rotations and three translations of each joint.

| Joint | Energy generated: 100% | | Energy absorbed: 100% | |
|---|---|---|---|---|
| | Stance: 88%(1%) | Swing: 12%(1%) | Stance: 56%(3%) | Swing: 44%(3%) |
| Elbow | 79(1%) | 52%(3%) | 61%(1%) | 98%(0%) |
| Carpus | 4%(0%) | 40%(3%) | 5%(0%) | 1%(0%) |
| Fetlock | 13%(1%) | 7%(1%) | 24%(1%) | 1%(0%) |
| Pastern | 1%(0%) | 0%(0%) | 3%(0%) | 0%(0%) |
| Coffin | 2%(0%) | 0%(0%) | 6%(0%) | 0%(0%) |